\documentclass{article} %
\usepackage[final]{corl_2025}
\usepackage{times}

\usepackage{amsmath,amsfonts,bm}

\def\eqref#1{equation~\ref{#1}}

\def\1{\bm{1}}

\DeclareMathAlphabet{\mathsfit}{\encodingdefault}{\sfdefault}{m}{sl}
\SetMathAlphabet{\mathsfit}{bold}{\encodingdefault}{\sfdefault}{bx}{n}

\usepackage{adjustbox}
\usepackage{algpseudocode}
\usepackage{algorithm}
\usepackage{amsfonts}
\usepackage{amsmath}
\usepackage{amssymb}
\usepackage{array}  %
\usepackage{authblk}
\usepackage{booktabs}
\usepackage{capt-of}
\usepackage[font={small}]{caption}
\usepackage{subcaption}
\usepackage{changepage}
\usepackage{colortbl}
\usepackage{rotating}
\usepackage{marginnote}
\usepackage[most]{tcolorbox}
\usepackage{epsfig}
\usepackage{etoolbox}
\usepackage{setspace}
\AtBeginEnvironment{quote}{\par\singlespacing\small}
\usepackage{float}
\usepackage{inconsolata}  %
\usepackage{graphicx}
\usepackage{listings}
\usepackage{multicol}
\usepackage{multirow}
\usepackage{ntheorem}
\usepackage{paralist}
\usepackage{pifont}
\usepackage{svg}
\usepackage{times}
\usepackage{tabularx}
\usepackage{thmtools}  %
\usepackage{wrapfig}
\usepackage{xcolor}

\newtheorem{hyp}{Hypothesis}

\newenvironment{SideBarBox}[3]{%
  \noindent
  \begin{tcolorbox}[
    enhanced,
    sharp corners,
    boxrule=0pt,
    colback=#2!8!white,
    frame hidden,
    left=#1,
    valign=top,
    top=0.8cm,
    bottom=0.8cm,
    overlay unbroken={%
      \path[fill=#2] ([xshift=0pt]interior.south west) rectangle ([xshift=#1]interior.north west);
      \node[rotate=90, font=\bfseries\footnotesize, text=white]
        at ([xshift=.5*#1]interior.west) {#3};
    }
  ]%
}{%
  \end{tcolorbox}\vspace{-2em}%
}

\newenvironment{PromptBlock}[1][]{%
  \begin{SideBarBox}{9mm}{orange}{Prompt}%
}{%
  \end{SideBarBox}%
}

\newenvironment{IntrospectionBlock}[1][]{%
  \begin{SideBarBox}{9mm}{purple}{Introspection}%
}{%
  \end{SideBarBox}%
}

\newenvironment{ActionBlock}[1][]{%
  \begin{SideBarBox}{9mm}{green!60!black}{Action}%
}{%
  \end{SideBarBox}%
}

\newenvironment{ReturnBlock}[1][]{%
  \begin{SideBarBox}{9mm}{gray}{Return}%
}{%
  \end{SideBarBox}%
}

\newenvironment{ToolCallBlock}[1][]{%
  \begin{SideBarBox}{9mm}{blue}{Tool Call}\ttfamily\small%
  
}{%
  \end{SideBarBox}%
}

\title{AutoODD: Agentic Audits via Bayesian Red Teaming in Black-Box Models}

\author{
  Rebecca Martin, Jay Patrikar, Sebastian Scherer\\
  Carnegie Mellon University, Field AI \\
  \texttt{\{rebeccacm, basti\}@cmu.edu, jay@fieldai.com} \\
}

\newcommand{\coolname}{\textit{AutoODD~}}

\begin{document}

\maketitle

\begin{abstract}
Specialized machine learning models, regardless of architecture and training, are susceptible to failures in deployment. With their increasing use in high risk situations, the ability to audit these models by determining their operational design domain (ODD) is crucial in ensuring safety and compliance. However, given the high-dimensional input spaces, this process often requires significant human resources and domain expertise. To alleviate this, we introduce \coolname, an LLM-Agent centric framework for automated generation of semantically relevant test cases to search for failure modes in specialized black-box models. By leveraging LLM-Agents as tool orchestrators, we aim to fit a uncertainty-aware failure distribution model on a learned text-embedding manifold by projecting the high-dimension input space to low-dimension text-embedding latent space. The LLM-Agent is tasked with iteratively building the failure landscape by leveraging tools for generating test-cases to probe the model-under-test (MUT) and recording the response. The agent also guides the search using tools to probe uncertainty estimate on the low dimensional manifold. We demonstrate this process in a simple case using models trained with missing digits on the MNIST dataset and in the real world setting of vision-based intruder detection for aerial vehicles. 
\end{abstract}

\section{Introduction}
\label{sec:intro}
\begin{figure*}[htb!]
\centering
\includegraphics[width=\linewidth]{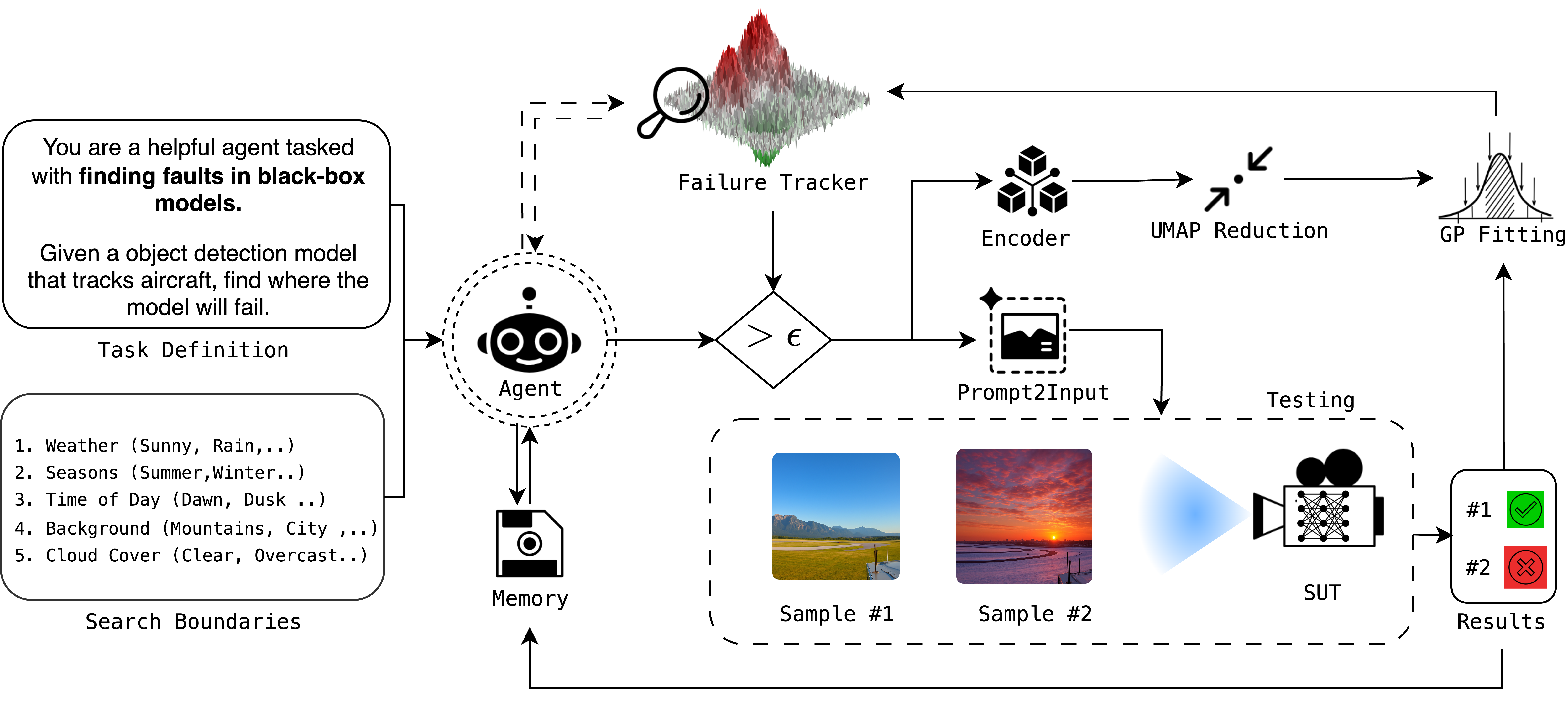}
\caption{The \coolname framework combines LLM-Agent orchestration with Bayesian uncertainty estimation to efficiently explore the failure landscape of black-box models in semantically meaningful embedding spaces.}
\label{fig:abstract}
\end{figure*}
Machine learning models are increasingly integrated into safety-critical robotic systems such as autonomous driving, surgical assistance, and industrial automation, where failures can have severe consequences. In these domains, it is essential to establish a model's \emph{operational design domain} (ODD)---the set of environmental and task conditions under which it can be expected to perform reliably \cite{wiesbrock2024blackbox}. For black-box models, where training data, architecture, and optimization details are unavailable, ODD estimation must be inferred empirically through systematic testing. This process is challenging: the input space is often high-dimensional, combinatorial in nature, and difficult to exhaustively explore, making naive random testing inefficient and prone to missing rare but critical failure modes.

A promising strategy is to project raw inputs (e.g., sensory data or scene descriptions) into a semantically meaningful embedding space, where related conditions are close together and the search space is dramatically reduced. Prior work shows that manifold-based active learning can improve sample efficiency by targeting diverse, informative regions of this space \cite{sreenivasaiah2021meal}. However, efficient exploration also requires an assessment of \emph{where} the model's behavior is uncertain. Bayesian optimization offers a principled mechanism for such exploration, using surrogate models such as Gaussian Processes (GPs) to estimate both performance and uncertainty, and to guide sampling toward high-value test cases \cite{lee2023bayesian}. This combination of semantic search spaces and uncertainty-guided sampling offers the potential to rapidly identify boundaries between safe and unsafe operation.

We introduce \textbf{AutoODD}, a framework for \emph{uncertainty-aware, agentic audits} of black-box models. AutoODD employs an LLM-based agent as a high-level orchestrator that interacts with the system-under-test via a set of tools: scenario generation, model querying, and uncertainty estimation. Given a task description and a semantically structured descriptor set, the agent proposes candidate scenarios in the embedding space. These are evaluated on the black-box model, with results used to update a GP surrogate that models performance over an embedding-manifold. The GP's uncertainty estimates interact with the agent's next generation step, creating a closed-loop \emph{generate$\rightarrow$test$\rightarrow$feedback} cycle \cite{danso2025llmods,xu2024llmtester}. By focusing on high-uncertainty or borderline cases, the framework systematically probes the edges of the model's competence with far fewer evaluations than unguided exhaustive search.

AutoODD's key contributions are twofold: \textbf{(1)} an agentic integration of semantic embedding spaces and Bayesian uncertainty estimation for black-box safety audits; and \textbf{(2)} a demonstration that this approach efficiently discovers operational boundaries and failure modes in both controlled and real-world safety-critical settings. We show that AutoODD can recover meaningful, human-interpretable failure landscape with significantly reduced sample complexity, providing a scalable and practical methodology for verifying the reliability of black-box models in safety-critical robotics.

\section{Related Work}
\label{sec:related}

\paragraph{Gaussian Surrogates for Failure Discovery}
Recent work applies Gaussian Processes (GPs) and Bayesian Optimization (BO) to systematically uncover critical scenarios in autonomous systems. One line of work models Signal Temporal Logic (STL) robustness with GPs and directs sampling toward low-robustness regions to efficiently falsify controllers \cite{torben2023}. Another applies BO directly in the semantic space of a promptable traffic simulator, inducing adversarial interactive driving behaviors that expose planner weaknesses \cite{mondelli2025}. Multi-objective extensions use multi-output GPs with negative-transfer mitigation to improve boundary test generation in multi-UAV coordination \cite{jiang2025}. In automotive settings, GP modeling combined with Sobol sensitivity analysis enables ranking of scenario parameters by their safety impact, allowing targeted exploration \cite{felix2023}. Manifold-informed BO explicitly constrains the search to low-dimensional subspaces for improved efficiency \cite{manifoldbo2023}, while active learning of GP embeddings improves sample allocation in high-dimensional search \cite{garnett2023}. Bayesian Red Teaming applies GP-BO to text embedding spaces with a diversity-penalized acquisition function to uncover varied adversarial prompts under strict budgets, a design transferable to physical testing \cite{brt2023}. Collectively, these studies confirm that GP-BO is sample-efficient and precise for boundary exploration, yet in high-dimensional spaces the posterior often highlights multiple “equally promising” directions, leaving the search vulnerable to local myopia without an external semantic driver.

\paragraph{LLM-Driven Critical Generation}
In parallel, large language models (LLMs) have been applied as flexible generators of diverse, realistic, and semantically grounded test scenarios. Pipelines have been introduced that translate natural language into validated Scenic/CARLA code, automatically producing executable safety-critical scenarios \cite{zhang2024}. Traffic-rule–guided approaches ground LLM outputs in domain-specific languages, reducing hallucination and generating verifiable rule-based scenarios \cite{target2023}. Feedback-driven generation–test–refine loops use multi-scale edits that combine coarse scene changes with fine perturbations to efficiently expose decision-making failures \cite{xu2024}. Retrieval and multimodal integration extend these ideas further: frameworks combine LLMs with traffic simulation and retrieval-augmented generation for high-fidelity multimodal scenarios \cite{omni2024}, while multimodal synthesis pipelines create realistic corner cases \cite{autoscn2024}. Adversarial methods edit risky agent trajectories to provoke collision-inducing events \cite{gao2025}, and retrieval-augmented pipelines remix real-world traffic data to build controllable synthetic tests \cite{ding2023}. Modular architectures ensure spatial consistency by grounding generation in retrieved road networks \cite{ruan2025}. These approaches excel at semantic novelty and diversity but lack the quantitative coverage tracking and uncertainty estimates provided by GP surrogates. A hybrid approach closes this gap by combining an LLM exploration branch for novelty and realism with a GP-guided exploitation branch for coverage tracking and targeted failure discovery, yielding both creativity and reliability.

\section{Approach}
\label{sec:methods}

\subsection{Background}

Consider a general setup where an LLM-agent $\mathcal{L}$ interacts with a system-under-test (SUT) $\mathcal{M}$. The SUT $\mathcal{M}(x) \rightarrow y$ takes as an input $x \in \mathcal{X}$ and outputs $y \in \mathcal{B}$ boolean which is true if SUT failed for $x$ and false if passed. Agent $\mathcal{L}$ at step $t$ can take an action $a_t \in \mathcal{A}$ and receive an observation $o_t \in \mathcal{O}$. In addition to this, the agent also can \textit{introspect} $i_t \in \mathcal{I}$. Given a templated text prompt $w$, let us also define a LLM encoder $\mathcal{E}(w) \rightarrow e$, where $e \in \mathbb{R}^n$ is the embedding for that prompt. Assume access to a generic \textit{prompt-to-input converter} $\mathcal{C}(w) \rightarrow x$ that maps the prompt $w$ to an input $x$ suitable for $\mathcal{M}$ (e.g., image, simulator configuration, or API call). In line with previous formulations on failure probability estimations \cite{moss2024bayesian}, let us also define $p(x)$ as the operational model and $f(x)$ define the failure model with respect to $\mathcal{M}$. To capture structure in the descriptor space, we fit \emph{one GP per axis} (category) over the embedding coordinates: let categories be indexed by $k=1,\dots,K$ with per-axis embedding components $e_k$, and maintain GP set $\{\mathcal{G}_k\}_{k=1}^K$ modeling axis-wise failure propensity and uncertainty. The action space $\mathcal{A}$ contains \textit{GenerateTest()}, which takes a templated prompt $w$, uses the converter $\mathcal{C}$ to construct an input $x$ for $\mathcal{M}$, queries $\mathcal{M}$, and returns both the generated input and outcome $(x, y)$.%

\subsection{Method}

\begin{algorithm}
\caption{\coolname algorithm}
\label{alg:cap}
\begin{algorithmic}
\While{$a_t \neq STOP$}
\State $a_t, w \gets \mathcal{L}$
\If{$a_t == GenerateTest()$}
    \State $r \gets \mathcal{U}(0,1)$
    \If{$r < \epsilon$} \Comment{exploit per GP posterior}
        \State $w \gets \arg\max_{w'} f(\mathcal{E}(w'))\,p(\mathcal{E}(w'))$
    \EndIf
    \State $x \gets \mathcal{C}(w)$
    \State $y \gets \mathcal{M}(x)$
    \State $e \gets \mathcal{E}(w)$
    \For{$k = 1$ to $K$}
        \State $\mathcal{G}_k.\text{fit}(e_k, y)$
    \EndFor
\EndIf
\EndWhile
\end{algorithmic}
\end{algorithm}

Figure \ref{fig:abstract} and algorithm \ref{alg:cap} provide an overview of the AutoODD approach. The LLM agent is provided with its task of finding failures, as well as a description of the problem that the SUT has been trained to solve and the function that it can use to interface with the black box model, namely \verb|generate_test()|. The prompt template is formed using a set of categories that form the axes for the search space, as well as all the keyword options for each category. The LLM agent then selects one keyword from each category to form one complete input description. We fit one GP per axis and use axis-wise uncertainty and mean estimates to guide both exploration and exploitation. When the black box model is queried, the text input is converted into a SUT-compatible input (e.g., image, simulator configuration), which is fed into the model, which in turn returns its performance for that input; \verb|generate_test()| returns both $(x,y)$ and the Gaussian processes are updated with this new information. For some $\epsilon$ fraction of steps, the agent \emph{exploits} by overriding its exploratory choice with the region of highest predicted failure likelihood under the posterior; otherwise, it explores based on semantic reasoning. After completion of the analysis, the LLM summarizes its findings in terms of the text keywords in a human understandable way.

\section{Experiments \& Results}
\label{sec:results}
We begin by testing the \coolname framework on MNIST with varying degrees of difficulty. We then demonstrate its efficacy in real world scenarios, using the Detect and Avoid problem for autonomous aircraft. 

\subsection{Experiment $\#$ 1: Colored MNIST}

The task is joint classification of digit (10-way) and color (10-way). We train a simple CNN with fixed hyperparameters across runs, varying only the training data curation.

We extend the standard MNIST dataset by coloring digits to introduce a second descriptor axis. The two axes are: (i) digit $\in {0..9}$ and (ii) color $\in$ {blue, orange, green, red, purple, brown, pink, gray, yellow, cyan}. Both axes are labeled outputs. Each digit–color pair is treated as a unique input, yielding 100 combinations.

We evaluate three training ablations:
\begin{itemize}
    \item Missing digit: remove all instances of the digit 7 (all colors) from training.
    \item Missing color: remove all instances of the color cyan (all digits) from training.
    \item Random sparse: remove 10 random digit–color pairs from training.
\end{itemize}

We then run \coolname to audit the trained CNN and discover the induced failure structure. The LLM interfaces with the detector via the \verb|generate_test()| function. Given the dataset’s small size, \verb|generate_test()| converts a (digit, color) query to an image by sampling a matching digit instance and applying the requested color at runtime; it returns both the constructed image and the model’s output.

\subsection{Experiment $\#$ 2: Detect and Avoid}

Detect and Avoid (DAA) is a crucial capability that aircraft require to avoid collisions with other airborne objects. The problem involves detecting, tracking, and maintaining safe separation from potential collision threats in three-dimensional airspace. DAA is particularly challenging due to its open-world, safety-critical nature. Aircraft operate across diverse environmental conditions including varying weather (rain, snow, fog), seasonal changes (lighting, visibility), cloud cover (clear to overcast), and different backgrounds (urban, mountainous, open terrain)\cite{tartanaviation}. While machine learning approaches can improve performance across these conditions, they may harbor hidden failure modes that only emerge under specific environmental combinations, making comprehensive safety validation essential. \cite{martin2024targeted}

In our experiments, we use AirTrack \cite{ghosh2022airtrack} as our black-box model. AirTrack consists of four sequential modules: Frame Alignment, Detection and Tracking, Secondary Classification, and Intruder State Update. The system takes two consecutive grayscale image frames as input and outputs a list of tracked objects with various attributes.

Unlike Experiment 1, this is a real-world system with no induced failures. The LLM interfaces with AirTrack through the \verb|generate_test()| function. For \verb|generate_test()|, we leverage GPT-4o to create images from templates based on the TartanAviation dataset \cite{tartanaviation}. These templates are then animated using aircraft tracks extracted from TartanAviation's static camera videos, enabling realistic scenario generation.

The LLM is provided with five categories: time of day (twilight, afternoon, sunset), cloud cover (clear, scattered, broken, overcast), background (city with low rises, city with high rises, small mountains, tall mountains, open fields, industrial area), season (summer, fall, winter with snow, rainy), and weather condition (none, heavy rain, snow, fog, thunderstorm). This yields 1440 possible input combinations. 

\subsection{Results \& Discussion}
\begin{figure}
    \centering
    \includegraphics[width=0.9\linewidth]{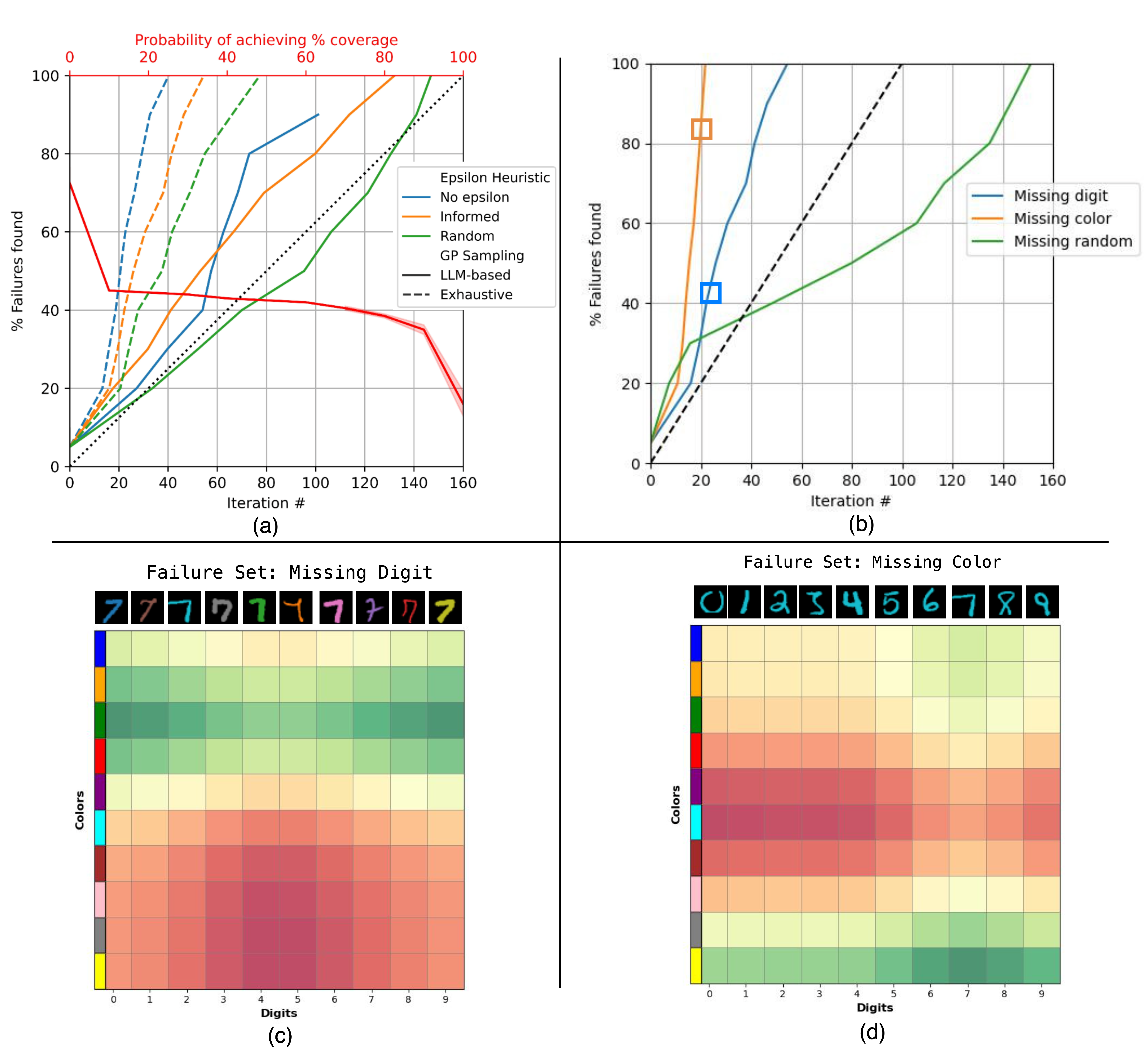}
    \caption{Fig 2a shows the failure discovery rate of the \coolname algorithm in the missing digit experiment with ablations for the epsilon heuristic and the Gaussian Process sampling. The black dotted line represents the expected performance of a random search as baseline. The red line shows the probability of achieving input space coverage with the LLM sampling the GP. Fig 2b shows the failure discovery rate with ablations for the failure case. Fig 2c and 2d are visualizations of the GP in the missing digit and color ablations respectively. }
    \label{fig:mnist_results}
\end{figure}

The failure cases of the MNIST model were created by removing all instances of an input from the training data. Out of the 100 possible inputs (all 10 digits in ten possible colors), we ran ablations with one digit missing, one color missing, and ten random inputs missing. All of the results can be seen in Figure \ref{fig:mnist_results}. 

\begin{hyp}[H\ref{hyp:first}] \label{hyp:first}
The \coolname algorithm performs best when there are patterns in the failures to be exploited.
\end{hyp}

In this first ablation study, Figure \ref{fig:mnist_results} shows that the LLM agent performs much better in the missing color and missing digit ablations, compared to the random failure case ablation. In these scenarios where there is a pattern in the failures, the LLM is able to discover, reason about, and exploit these patterns. Comparing between the two, the missing color case performs the best as the input embeddings for the digits were closer to each other in latent space than the embeddings of the colors, which can be mitigated by normalizing the Gaussian Process. %

\begin{hyp}[H\ref{hyp:second}] \label{hyp:second}
The \coolname algorithm guarantees coverage of the defined input space with $\epsilon \neq 0$, but will prioritize exploitation when $\epsilon = 0$ \end{hyp}

The second set of ablations focuses on the epsilon heuristic. The epsilon heuristic samples the Gaussian process and overrides the LLM query every epsilon percent of the time. We tested out epsilon values of 0 and 0.1, to test the efficacy of this heuristic. While the overall analysis finishes much faster with an epsilon of 0.0, this system does not provide any guarantees of input coverage. Having a higher epsilon value allows the algorithm to better exploit any patterns in the failures, but its lack of exploration leaves it susceptible to missing any disjoint failures. This can be accounted for by changing the epsilon heuristic from choosing the maximum uncertainty input to a random input, but this will decrease the exploitation benefit, as shown in the figure.

\begin{hyp}[H\ref{hyp:third}] \label{hyp:third}
The benefit of the Gaussian Process increases with the size of the input space.  \end{hyp}

\begin{figure}
    \centering
    \includegraphics[width=0.95\linewidth]{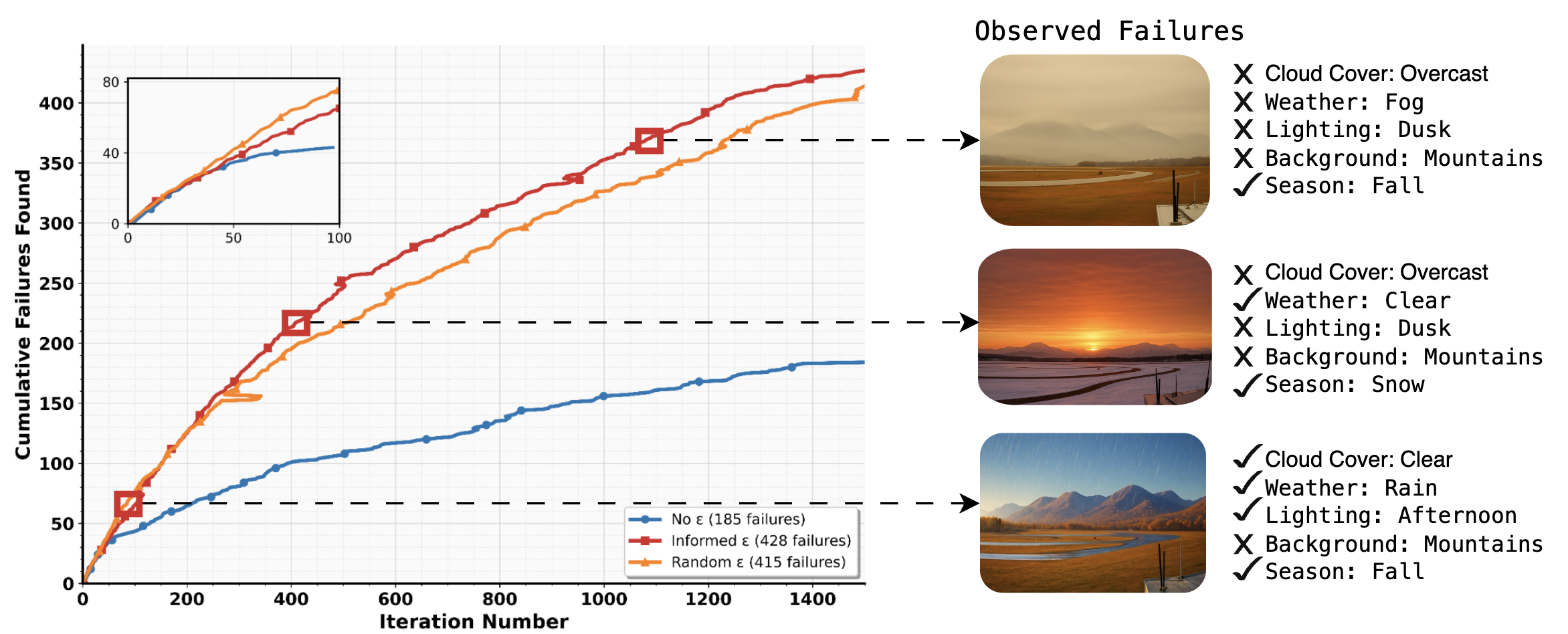}
    \caption{The plot shows the results of the \coolname framework on the real world Detect and Avoid use case, based on how quickly it balances exploiting failure modes with exploring the input space. On the right are examples of failure inputs that the framework has discovered, using the failure patterns to guide its search. }
    \label{fig:daa_summary}
\end{figure}

As the size of the input space increases, the Gaussian Process becomes more important in keeping track of which inputs have been queried and what their results are. For smaller input spaces, the informed and random heuristic perform similarly, due to the Gaussian Process being fit on a small number of points. However, the benefit of the informed heuristic shows through for more complex problems with larger input spaces, as seen in Figure \ref{fig:daa_summary}.

Overall, the results in the Detect and Avoid setting are consistent with the findings from the MNIST ablations studies. The overall performance summary as given by the LLM is shown in Figure \ref{fig:daa_summary}.

We also performed one ablation in the real world case, testing the different heuristics. Consistent with the earlier ablations, we see that the informed epsilon case is able to discover failures faster than a random sampling heuristic. It also finds many more failures when compared to removing the epsilon case, because it places more focus on the exploration component, in the case of disjoint failure modes. 

\begin{hyp}[H\ref{hyp:fourth}] \label{hyp:fourth}
Adding the Gaussian Process explicitly to the LLM reasoning performs worse than \coolname, which overrides with epsilon  \end{hyp}

In the $\epsilon = 0$ case, we let the LLM sample the Gaussian Process directly using a \verb|get_uncertainty()| function, which takes a prompt $w$ and uses the embedding model $\mathcal{E}$ with the GP surrogates to provide a failure-uncertainty estimate $u$ for that prompt. However, this underperforms both heuristics significantly, even though the frequency of sampling is not fixed and therefore ends up being much oftener than the heuristic sampling. Two contributing factors to this are the variability in choosing which inputs to sample and the possibility of choosing a repeat input.

\section{Conclusion}
\label{sec:conc}

In summary, this papers introduces AutoODD, a framework for uncertainty-aware, agentic audits of black-box models. AutoODD employs an LLM-based agent as a high-level orchestrator that interacts with the system-under-test via a set of tools: scenario generation, model querying, and uncertainty estimation. AutoODD’s key contributions are a novel agentic integration of semantic embedding spaces and Bayesian uncertainty estimation for black-box safety audits and a demonstration that this approach efficiently discovers operational boundaries and failure modes in both controlled and real-world robotics settings. AutoODD is able to recover a meaningful, human-interpretable failure landscape with significantly reduced sample complexity, providing a scalable and practical methodology for verifying the reliability of black-box models in safety-critical robotics. In the future, we will consider an open vocabulary problem formulation, allowing for a more generalizable framework and removing the need for a human expert to define the input space.

\bibliography{ref}

\section{Appendix}
\appendix
\section{LLM Agentic Conversation Example}
\label{sec:appendix_conversation}

This appendix demonstrates the interactive conversation between the LLM agent and the AutoODD framework during the Detect and Avoid (DAA) experiment. The conversation showcases the agent's introspection, tool usage, and iterative failure discovery process.

\subsection{Example Conversation}

\begin{PromptBlock}
You are given access to a model that reads digits in images, and your goal is to find which color and digit combinations the model performs poorly on.
Each digit (0-9) can appear in one of ten colors:
blue, orange, green, red, purple, brown, pink, gray, yellow, and cyan.
Solve this task by alternating between:
Introspect: Reason about what you've learned so far and decide your next step.
Action: Call exactly one function in each Action step. 
\\ --- \\
Available Actions:
You may use the following two functions:
\begin{enumerate}
    \item generate(color, digit)
    \begin{itemize}
        \item Tests the model on the specified digit in the given color
        \item Returns 1 for success (correct prediction) and 0 for failure
        \item The results of this function are deterministic and WILL NOT CHANGE if queried multiple times
        \item Example: generate('purple', '6')
        \item After each generate call, you will be given the result and asked if you want to query more or if you are ready to give a summary
        \item DO NOT REPEAT generate calls for the same (color, digit) pair
    \end{itemize}
    \item STOP(summary)
    \begin{itemize}
        \item Use this function when you are finished querying the model and output a summary of your findings
        \item You can only call STOP() directly after a generate call
    \end{itemize}
\end{enumerate} 
--- \\
Rules
\begin{itemize}
    \item Every response must contain one Introspect step and one Action step and NOTHING ELSE
    \item DO NOT REPEAT generate calls for the same (color, digit) pair, the result WILL NOT change
    \item Only one function call is allowed per Action step
    \item After every generate call, you will be given the result and asked if you want to query more or if you are ready to give a summary
\end{itemize}
--- \\
Task Completion Rules:
\begin{itemize}
    \item You must call generate() on all (color, digit) pairs you believe may fail
    \item Do not stop until all failure-prone inputs have been directly tested with generate()
    \item Do not assume a pattern (e.g., “all purples fail”) without confirming each suspected input
    \item Never repeat function calls for the same (color, digit) pair
\end{itemize}
--- \\
Notes for Introspection

When summarizing patterns, remember:
\begin{itemize}
    \item If a pattern suggests a full row or column of failures (e.g., all 6s or all purples), you still need to test each of those combinations with generate() to support your claim
    \item If uncertainty is high for a combination, you must confirm the failure with a generate() call before using it in your final summary
\end{itemize}
\end{PromptBlock}

\begin{IntrospectionBlock}

To assess the performance of the airplane detection model across different environmental conditions, I need to evaluate which specific combination of conditions leads to poor model performance (resulting in a 0).

From the previous function calls and results, it appears that the model performs poorly in very specific conditions, especially under heavy rain and fog. However, some combinations with clearer weather did yield better results.

Considering this information, I will extract the categories with the most failed performance and continue testing:

\begin{enumerate}
    \item **Time of Day**: Twilight
    \item **Sky Cover**: Overcast
    \item **Background**: Industrial area
    \item **Season**: Summer
    \item **Weather Condition**: Heavy rain
\end{enumerate}

I will query this combination to further establish the model's limits in this specific case.

\end{IntrospectionBlock}

\begin{ActionBlock}
Let's run the first query to examine how it performs under the identified condition:
\end{ActionBlock}

\begin{ToolCallBlock}
generate('twilight', 'overcast', 'industrial area', 'summer', 'heavy rain')
\begin{center}
  \vspace{0.2cm}
  \includegraphics[width=0.7\textwidth]{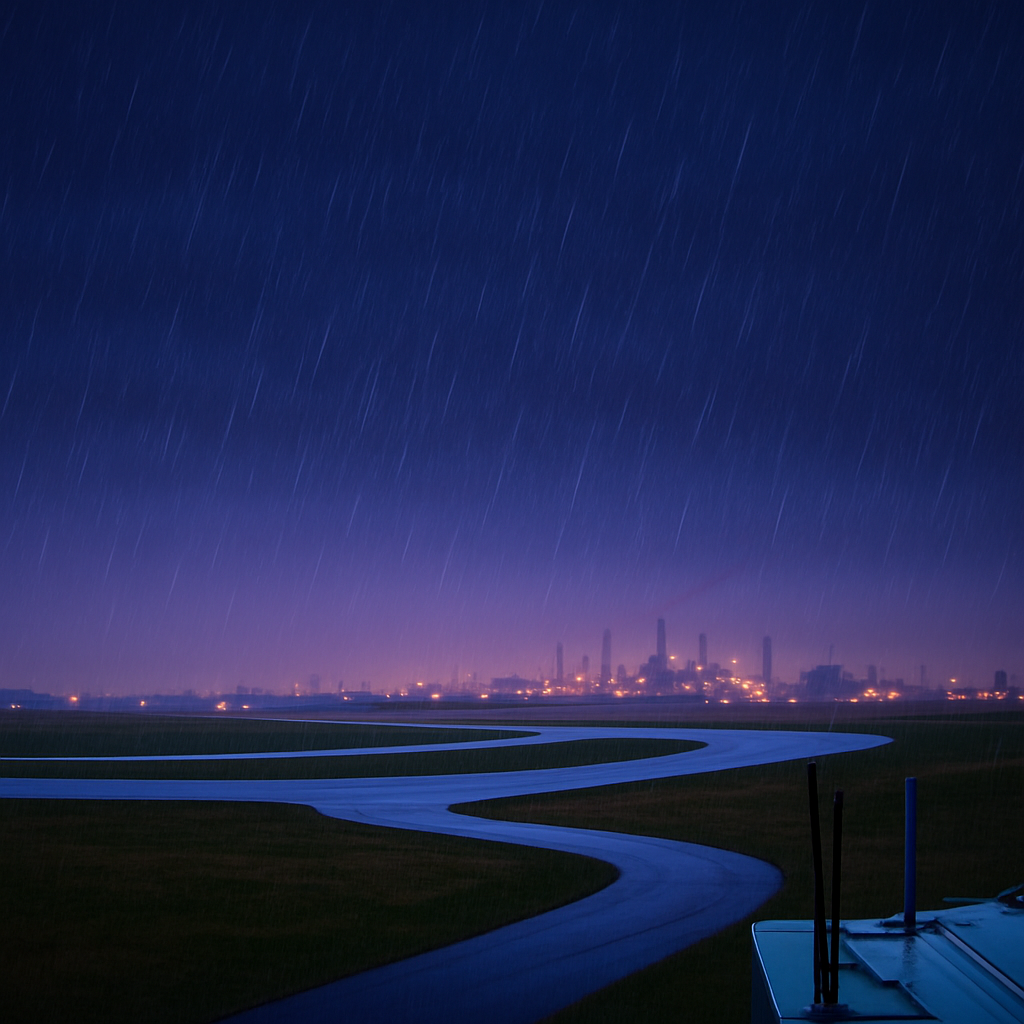}
  \vspace{0.2cm}
  \end{center}

\end{ToolCallBlock}

\begin{ReturnBlock}
\textbf{Result:} 0 $\rightarrow$ The model failed.
\end{ReturnBlock}

\end{document}